\documentclass[letterpaper, 10 pt, conference]{ieeeconf}
\IEEEoverridecommandlockouts
\usepackage{graphics}
\usepackage{graphicx}
\usepackage{epsfig}
\usepackage{mathptmx}
\usepackage{amsmath}
\usepackage{amssymb}
\DeclareMathAlphabet{\mathcal}{OMS}{cmsy}{m}{n}
\usepackage{cite}
\usepackage{url}
\usepackage{booktabs}
\usepackage{multirow}
\usepackage{makecell}
\usepackage{colortbl}
\usepackage{cuted}
\usepackage{capt-of}
\usepackage{algorithm}
\usepackage[hidelinks]{hyperref}
\usepackage{paralist}
\usepackage{algpseudocode}
\usepackage{placeins}
\newlength{\redactwidth}

\newsavebox{\identitybox}

\renewcommand{\paragraph}[1]{\par\indent\textbf{#1.}\enspace\ignorespaces}

\title{\LARGE \bf REDACT: Robust Perceptive Locomotion under Unseen \\[2pt] Visual Corruption}

\hypersetup{
    pdftitle={REDACT: Robust Perceptive Locomotion under Unseen Visual Corruption},
    pdfauthor={Natapat Kirdwichai, Tobias Driskell-Poole, Andrei Sontea, Jadu Dash, Muhammad Burhan Hafez, Danesh Tarapore}
}

\author{Natapat Kirdwichai, Tobias Driskell-Poole, Andrei Sontea, Jadu Dash, \\[2pt] Muhammad Burhan Hafez, and Danesh Tarapore
\thanks{The authors are with the Faculty of Engineering and Physical Sciences and the Faculty of Environmental and Life Sciences, University of Southampton, United Kingdom. Email: {\tt\small nk3g22@soton.ac.uk}}}

\begin{document}

\bstctlcite{IEEEreferencecontrol}
\maketitle
\thispagestyle{empty}
\pagestyle{empty}

\begin{abstract}
    Depth-conditioned locomotion policies have demonstrated impressive agile maneuvers, but can be steered to unpredictable actions when observations are outside their training distribution. Occlusion, invalid returns, sensor noise, and visual distractors can shift deployment observations away from nominal simulated depth. While synthetic sensor augmentation targets specified degradations, it does not by itself define behavior under corruption families omitted from training. To address gaps in training-time coverage, we present REDACT (Retaining Evidence Despite Artifacts for Continued Traversal), a teacher--student framework combining an improved visual encoder architecture, persistent feature masking, and a novel consensus-gating algorithm to retain useful depth information under unmodeled corruption. The gate uses approximate conformal calibration on clean observations alone, requiring no prior knowledge of the corruption type. Trained on clean simulated depth, REDACT retains useful visual information under unseen corruption, supporting higher traversal success than existing parkour baselines. Evaluation of depth augmentation across corruption families further shows that REDACT improves robustness where augmentation coverage is missing. Real-world trials demonstrate zero-shot transfer to structured and forested environments with unfamiliar scene content. Project page: \url{https://gatjungk.github.io/REDACT/}.

\end{abstract}

\suppressfloats[t]
\begin{figure}[t]
    \centering
    \includegraphics[width=\columnwidth]{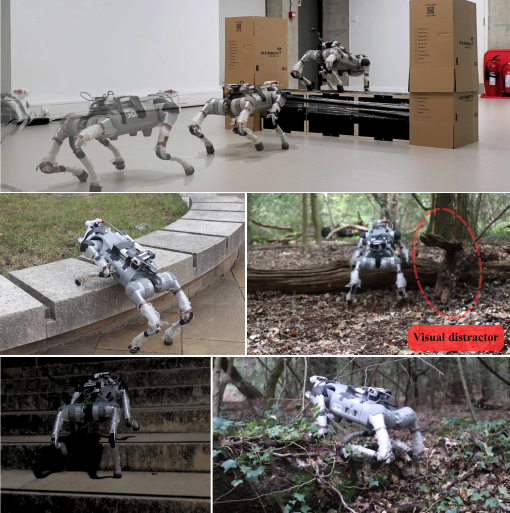}
    \caption{Zero-shot real-world traversal with REDACT. Repeated trials cover 0.3~m hurdles with out-of-distribution adjacent pillars (top), a log with fallen-tree distractors (middle right), and a log amid vegetation and leaf litter (bottom right). Depth dropouts and partial tree occlusion in this setting are illustrated in Fig.~\ref{fig:corruptions}. Additional panels demonstrate successful traversal of structured outdoor terrain and climbing steps under low ambient lighting.}
    \label{fig:real_world_teaser}
\end{figure}

\section{Introduction}
\label{sec:introduction}

Forest environments bring together many of the challenges faced by ground robots operating outside structured settings. Roots, deadwood, vegetation, and abrupt elevation changes interrupt traversal and limit autonomous coverage~\cite{kirdwichai2026forent}. Legged robots offer a range of movements for negotiating such terrain, including jumps, climbs, and coordinated changes in body posture. These maneuvers require the robot to interpret terrain information to anticipate suitable footholds and adjust body motion before contact.

Deep reinforcement learning has enabled these agile maneuvers through policies that combine terrain observations with proprioceptive feedback~\cite{miki2022wild,agarwal2022egocentric,cheng2023extreme}. Training in massively parallel simulation exposes policies to diverse terrain configurations and physical conditions, supporting robust locomotion and sim-to-real transfer~\cite{rudin2022minutes}. Simulated sensor augmentation extends this randomization to perceptual inputs by introducing degraded observations during training. This exposure supports learning both terrain representations that tolerate sensor errors and mechanisms that regulate reliance on vision through gating and multimodal integration~\cite{miki2022wild}.

Despite these advances, reproducing the full range of visual variation encountered in unstructured environments remains difficult. While egocentric depth policies avoid geometric information loss associated with elevation-map construction, their richer observations also expose the policy to greater variation in scene content and sensor artifacts~\cite{agarwal2022egocentric,rudin2025wild}. In forests, diverse spatial arrangements of trees and ground clutter, occlusion from overhanging vegetation, and weather-dependent sensor responses produce observation changes that are difficult to capture fully during training. Without corresponding training examples, the encoder may fail to preserve useful terrain information under unfamiliar visual changes. At deployment, out-of-distribution observations may therefore distort the terrain representation and induce destabilizing control actions, leading to collisions or falls.
 
To address gaps in training-time visual coverage, we introduce \emph{REDACT} (Retaining Evidence Despite Artifacts for Continued Traversal). Rather than requiring depth augmentation for each anticipated corruption, REDACT combines architectural advances beyond the shallow visual encoders common in perceptive locomotion with a novel consensus-gating algorithm to reduce sensitivity to perturbations and retain useful terrain information for agile locomotion under unseen visual degradation. The gate selectively aggregates spatial features based on disagreement among local predictions of a shared representation, using thresholds set through approximate conformal calibration on clean observations alone. This calibration requires neither corruption examples nor specification of the corruption family. Our experiments show REDACT’s improved robustness to corruptions absent from training, with further gains from augmentation.

Our main contributions are summarized as follows:
\begin{itemize}
    \item We propose REDACT, a depth-conditioned locomotion framework combining improved visual encoding, persistent feature masking, and a novel consensus-gating algorithm to retain useful terrain information for agile locomotion under corruption absent from training.

    \item Our consensus gate identifies inconsistent spatial features from disagreement among local predictions of shared visual features, with per-cell thresholds calibrated on clean rollouts alone, requiring neither corruption examples nor specification of the corruption family.

    \item We assess the generalizability of depth augmentation on a representative baseline architecture~\cite{cheng2023extreme}, demonstrating uneven transfer to unseen corruptions. These comparisons show REDACT's utility in addressing coverage gaps: without augmentation, REDACT achieves 74\% mean success versus 37\% for augmented baselines on corruptions absent from their training at maximum tested severity, with further gains from augmentation.

    \item Real-world trials demonstrate REDACT's zero-shot traversal over structured obstacles and forest terrain with unfamiliar visual content.
\end{itemize}

\section{Related Work}
\label{sec:related_work}
\suppressfloats[t]
\begin{figure}[t]
    \centering
    \includegraphics[width=\columnwidth,trim=0 2bp 0 2bp,clip]{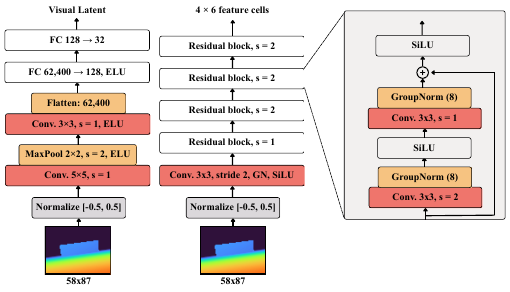}
    \caption{Representative ConvNet architecture~\cite{cheng2023extreme} (left; approximately 8 million parameters) and REDACT's residual encoder before gating and pooling (right; approximately 1.5 million parameters). Labels specify stride $s$; the inset shows a stride-2 residual block. GN denotes group normalization.}
    \label{fig:residual_encoder}
\end{figure}

\begin{figure*}[t]
    \centering
    \includegraphics[width=0.99\textwidth]{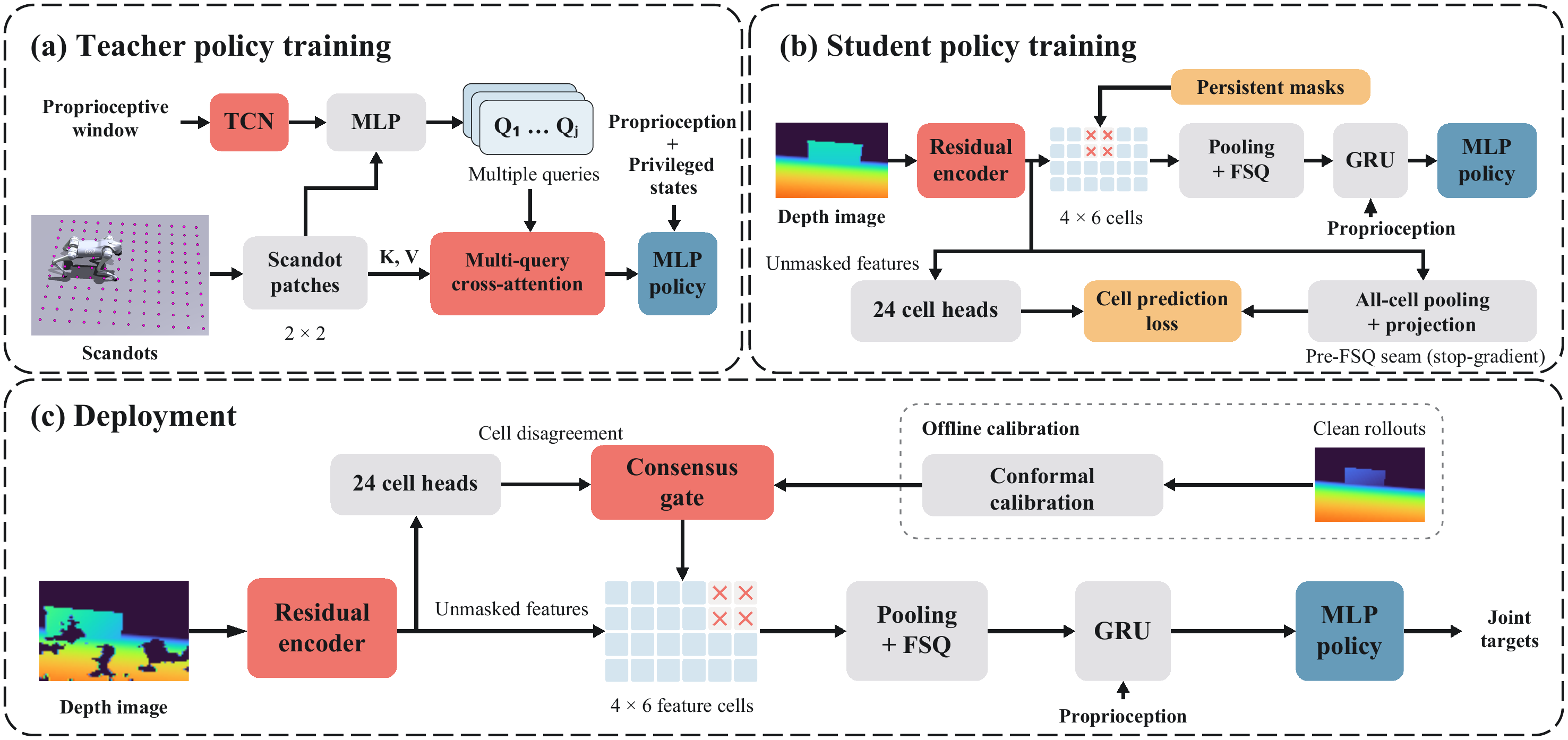}
    \caption{REDACT training and deployment. (a) A multi-query privileged teacher initializes the student actor. (b) Persistent masks train control from partial evidence; 24 cell heads predict the unmasked pre-FSQ latent with a stop-gradient target. (c) Clean-rollout calibration sets gating thresholds; retained features are pooled and quantized for recurrent control. Pathways of teacher supervision and proprioception to the student MLP policy are omitted for clarity.}
    \label{fig:redact_overview}
\end{figure*}

Transferring depth-conditioned locomotion policies to real environments requires accounting for discrepancies between simulated and measured depth. Synthetic sensor augmentation exposes policies to anticipated degradations during training, while preprocessing, such as downsampling and Gaussian blurring~\cite{rudin2025wild}, brings simulated and real observations closer in appearance by reducing fine-scale differences. Successful deployment in unstructured outdoor environments supports their combined use for locomotion despite imperfect depth sensing~\cite{rudin2025wild,agarwal2022egocentric,li2026kivi}. However, these results do not establish whether robustness extends to unmodeled corruptions arising under different environmental and weather conditions.

\paragraph{Perceptual Gating}
Alongside improving tolerance to degraded observations, prior work has shown that regulating perceptual contributions and relying on proprioception and memory can sustain locomotion under visual degradation. For example, REAL uses proprioception-conditioned feature-wise linear modulation (FiLM) to adapt visual features during motion~\cite{liu2026real}, while KiVi separates body-state encoding from memory-based terrain estimation to support control when vision is unreliable~\cite{li2026kivi}. RENet similarly supports proprioceptive fallback, using depth reconstruction error to detect deviations from its learned image distribution and switch to a proprioceptive estimator~\cite{zhang2025renet}. This avoids reliance on degraded visual estimates but discards useful obstacle information when only part of the observation is corrupted. Miki et al.~\cite{miki2022wild} demonstrate a more selective approach through a recurrent belief encoder whose gate regulates exteroceptive contributions. The gate is trained with hand-designed perturbations of terrain height samples to represent mapping errors and perception failures. However, extending such learned selection to egocentric depth requires retaining useful visual information under diverse scene-dependent degradations, including those not represented during gate training.

\paragraph{Robust Aggregation and Visual Encoding}
Selective rejection has also been studied in image classification, where corrupted spatial features can be removed before they influence the final prediction. PatchGuard limits the number of features affected by a localized adversarial patch using small receptive fields, then masks suspicious regions before aggregation~\cite{xiang2021patchguard}. Its robustness guarantees allow arbitrary corruption within a patch but require a bound on its spatial extent. In locomotion, sensor noise, occlusion, and unfamiliar scene geometry can affect depth observations over varying spatial extents, making such a bound difficult to specify.

Beyond gating under degraded observations, perceptual robustness also depends on how visual information is encoded. Depth-conditioned locomotion methods commonly use shallow convolutional encoders (e.g., Fig.~\ref{fig:residual_encoder}, left) to provide visual features for terrain estimation, world modeling, and temporal fusion~\cite{cheng2023extreme,luo2024pie,lai2025wmp,liu2026real}. Therefore, errors introduced during visual encoding can propagate to these downstream components. In broader visual reinforcement learning, IMPALA demonstrated improved policy performance with a deep residual encoder~\cite{espeholt2018impala}. Building on this architecture, Beyond the Rainbow improved learning performance through architectural changes including spectral normalization and adaptive max pooling~\cite{clark2025btr}. Whether these encoder improvements also confer robustness to depth corruptions absent from training remains open in perceptive locomotion.

\section{Method}
\label{sec:method}

We train a privileged teacher and a depth-conditioned student to retain terrain information needed for locomotion while limiting the influence of unseen visual corruption (Fig.~\ref{fig:redact_overview}). Regularized multi-query cross-attention first structures the teacher's terrain representation to reduce dependence on privileged detail that onboard depth cannot recover. The teacher then initializes a depth-conditioned student whose residual encoder is designed to limit sensitivity to visual perturbations while preserving spatial evidence for consensus gating. Persistent masking prepares the controller to act on partial feature grids, while clean-calibrated thresholds guide feature removal at deployment. Additional implementation details and hyperparameters will be provided on the project page.

\subsection{Teacher Training}
\label{sec:teacher_encoder}

We extend the single-query cross-attention used in prior work~\cite{liu2026real,zhang2026ame2} to multiple queries, and regularize the attention and resulting terrain representation. This aims to support fine-tuning of the student actor initialized with the teacher's weights~\cite{cheng2023extreme} by reducing dependencies on privileged information that onboard depth cannot fully recover  (Fig.~\ref{fig:redact_overview}(a)). To form the queries $q_j$, an MLP fuses proprioceptive history encoded by a temporal convolutional network (TCN) with pooled scandot features. Keys $K$ and values $V$ come from $2\times2$ scandot patches, with each query retrieving
\begin{equation}
    a_j=\operatorname{softmax}\!\left(\frac{q_j K^{\mathsf T}}{\sqrt{d_k}}\right),
    \qquad r_j=a_jV,
    \label{eq:teacher_attention}
\end{equation}
where $a_j$ contains attention weights, $r_j$ is the retrieved feature, $j=1,\ldots,4$, and $d_k=64$ is the query/key dimension. Separate retrievals allow the teacher to represent terrain regions jointly relevant to a maneuver, such as takeoff and landing surfaces. To concentrate these retrievals rather than distribute attention broadly across terrain, we penalize attention entropy, while an overlap penalty encourages different queries to attend to distinct regions~\cite{lin2017structured}:
\begin{equation}
    \mathcal L_{\mathrm{ent}}
        =-\frac{\mathbb E_j[\sum_n a_{j,n}\ln a_{j,n}]}{\ln N},
    \quad \mathcal L_{\mathrm{ov}}
        =\mathbb E_{j\ne k}[a_j a_k^{\mathsf T}].
    \label{eq:teacher_regularizers}
\end{equation}
Here, $n$ indexes $N$ scandot patches, and both penalties are averaged over the training batch. The retrieved features are concatenated and projected into a tanh-bounded terrain latent $z_t^{\mathrm T}\in\mathbb R^{32}$, which conditions the actor alongside proprioception and privileged states.

To learn agile locomotion, we train the teacher using proximal policy optimization (PPO)~\cite{schulman2017ppo}, with the reward formulation from~\cite{cheng2023extreme} and separately weighted regularizers in Eqs.~\eqref{eq:teacher_regularizers} and~\eqref{eq:teacher_slowness}. To make the terrain representation more predictable for the student, we further penalize changes between consecutive observations within a trajectory:
\begin{equation}
    \mathcal L_{\mathrm{slow}}
        =\frac{\mathbb E[\|z_{t+1}^{\mathrm T}-z_t^{\mathrm T}\|_2^2]}{d_z}.
    \label{eq:teacher_slowness}
\end{equation}
Here, $d_z=32$ is the latent dimension, and the expectation is over consecutive observation pairs.

\subsection{Student Training}
\label{sec:spatial_encoder}

The student replaces privileged terrain sensing with a spatial depth representation (Fig.~\ref{fig:redact_overview}(b)). To improve robustness to depth perturbations, we adopt an IMPALA-style residual architecture~\cite{espeholt2018impala} in place of the shallow ConvNet commonly used in depth-conditioned locomotion. Group normalization~\cite{wu2018groupnorm} is incorporated to regulate intermediate feature scales, while strided convolutions replace intermediate max pooling for downsampling. The encoder maps each depth image to a $4\times6$ grid of feature cells $f_\ell$ for selective aggregation through consensus gating (Fig.~\ref{fig:residual_encoder}, right). For a retained subset $S$, mean--max pooling and a spectrally capped projection~\cite{clark2025btr} form the continuous visual latent
\begin{equation}
    u(S)=\tanh\!\left(
    \operatorname{Proj}\!\left[
    \frac{1}{|S|}\sum_{\ell\in S}f_\ell,\;
    \max_{\ell\in S}f_\ell
    \right]\right).
    \label{eq:masked_latent}
\end{equation}
We use finite scalar quantization (FSQ)~\cite{mentzer2024fsq} to map each latent coordinate to five levels, leaving the visual code unchanged when perturbations from cell removal or input noise remain within their quantization bins. The quantized latent and proprioception enter a gated recurrent unit (GRU) that supplies heading errors and a learned representation to the initialized actor, replacing the privileged scandot latent.

\begin{algorithm}[t]
    \caption{Consensus-gated visual update}
    \label{alg:consensus_gate}
    \small
    \algtext*{EndIf}
    \begin{algorithmic}[1]
        \Statex \textbf{Inputs:} Depth $d_t$, proprioception $p_t$, recurrent state $h_{t-1}$, calibrated thresholds $\tau_\ell$.
        \State $\{f_\ell\}_{\ell=1}^{24}\gets\operatorname{Encoder}(d_t)$
        \State $\hat u_\ell\gets\tanh(W_\ell f_\ell+b_\ell),\quad \ell=1,\ldots,24$
        \State $s_\ell\gets\|\hat u_\ell-\operatorname*{med}_{j\ne\ell}\hat u_j\|_2,\quad \ell=1,\ldots,24$
        \State $S\gets\{\ell:s_\ell\le\tau_\ell\}$
        \If{$|S|<6$}
            \State Sort cell indices $\ell_1,\ldots,\ell_{24}$ by increasing score
            \State $S\gets\{\ell_1,\ldots,\ell_{18}\}$
        \EndIf
        \State $z_t\gets\operatorname{FSQ}(u(S))$ \Comment{$u(S)$: Eq.~\eqref{eq:masked_latent}}
        \State $h_t\gets\operatorname{GRU}([z_t,p_t],h_{t-1})$
    \end{algorithmic}
\end{algorithm}

% Removed "with a tanh output" to reduce space in the paragraph below

To limit corruption before quantization and recurrent processing, REDACT uses disagreement between cell predictions to identify inconsistent feature contributions. Unseen depth structures can affect spatial features unevenly. By training each cell to predict the same unmasked full-map latent, we establish a common target for comparing their predictions despite differences in local image content. A location-specific linear head produces $\hat u_\ell$, supervised by
\begin{equation}
    \mathcal L_{\mathrm{cell}}
    =\frac{1}{24D}\sum_{\ell=1}^{24}
    \left\|\hat u_\ell-\operatorname{sg}[u(I)]\right\|_2^2,
    \label{eq:cell_loss}
\end{equation}
\suppressfloats[t]
\begin{figure}[!t]

    \centering
    \includegraphics[width=\columnwidth]{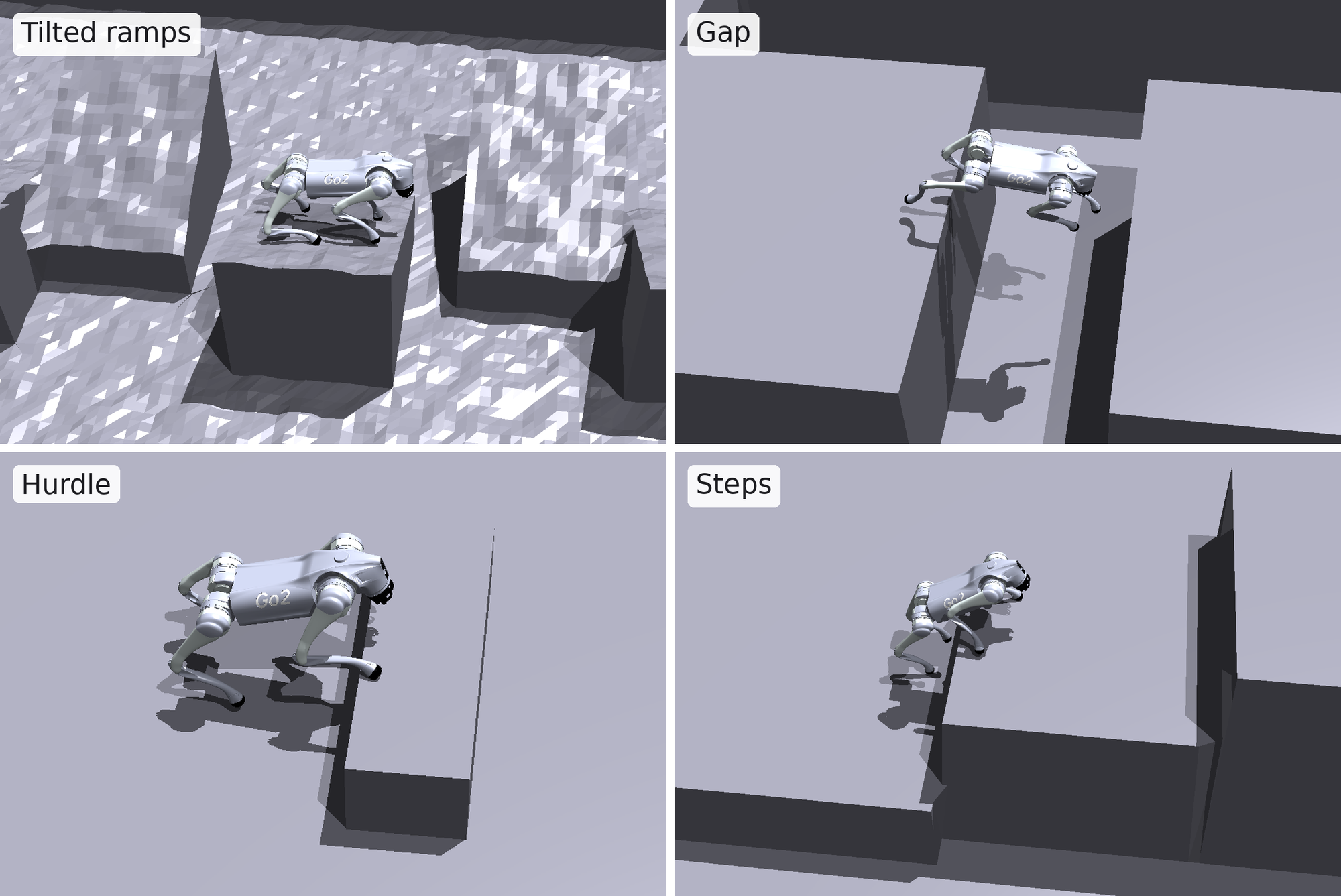}
    \caption{Simulated training terrains: tilted ramps, gaps, hurdles, and steps.}
    \label{fig:terrains}
    \par\medskip
    \input{Figures/robustness_float}
\end{figure}

\begin{figure*}[t]
    \centering
    \includegraphics[width=\textwidth]{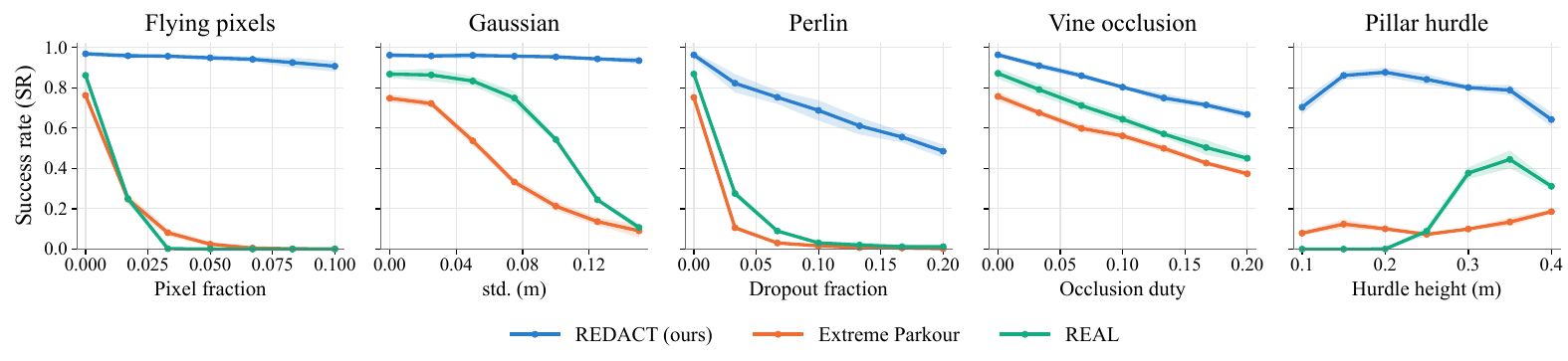}
    \caption{Success rate across four depth-corruption sweeps and pillar-hurdle trials. Lines show means and shaded bands indicate 95\% confidence intervals. For pillar hurdles, hurdle and pillar heights vary together, with pillars three times the hurdle height, rather than varying image-corruption severity.}
    \label{fig:baseline_comparison}
\end{figure*}

where $I$ contains all cells, $D=64$, and $\operatorname{sg}$ denotes stop-gradient. At deployment, consensus gating uses this disagreement to exclude inconsistent cells from pooling, limiting their influence on the visual representation supplied to the recurrent policy. To discourage latent collapse, we penalize feature-wise batch standard deviations below $\gamma=0.22$~\cite{bardes2022vicreg}:
\begin{equation}
    \mathcal L_{\mathrm{var}}
    =\frac{1}{D}\sum_{k=1}^{D}
    \max\!\left(0,\gamma-\operatorname{Std}[u_k(I)]\right).
    \label{eq:variance_loss}
\end{equation}

We train with intermittent feature masks to prepare the recurrent policy for partial visual evidence, including information removed by gating. Contiguous masks following DropBlock~\cite{ghiasi2018dropblock} and scattered cell removal are held fixed across several camera updates. Masks are sampled independently of consensus scores and applied only to pooling, leaving the cell heads and their full-map supervision unchanged.

The student collects trajectories under its own policy, with the teacher supervising the visited states through DAgger-style behavioral cloning~\cite{cheng2023extreme}. The visual encoder, GRU, and initialized actor are optimized jointly:
\begin{equation}
    \mathcal L_{\mathrm S}
    =\mathbb E\!
    \bigl(
    \lVert a^{\mathrm S}-a^{\mathrm T}\rVert_2
    +
    \lVert\hat\psi-\psi^{\mathrm T}\rVert_2
    \bigr)
    +\lambda_c\mathcal L_{\mathrm{cell}}
    +\lambda_v\mathcal L_{\mathrm{var}},
    \label{eq:student_loss}
\end{equation}
where $a^{\mathrm S}$ and $a^{\mathrm T}$ are student and teacher actions, and $\hat\psi$ and $\psi^{\mathrm T}$ are predicted and target heading errors. Student predictions use the sampled pooling subset, and $\lambda_c$ and $\lambda_v$ weight the auxiliary losses.

\suppressfloats[t]
\begin{table}[t]
    \centering
    \caption{Evaluation conditions and severity ranges.}
    \label{tab:corruptions}
    \footnotesize
    \setlength{\tabcolsep}{3pt}
    \renewcommand{\arraystretch}{1.2}
    \begin{tabular}{@{}>{\raggedright\arraybackslash}p{0.20\columnwidth}>{\raggedright\arraybackslash}p{0.22\columnwidth}>{\raggedright\arraybackslash}p{\dimexpr0.58\columnwidth-12pt\relax}@{}}
        \toprule
        \textbf{Condition} & \textbf{Severity range} & \textbf{Description} \\
        \midrule
        Flying pixels & 0--10\% of pixels & Set randomly selected pixels to depths sampled uniformly from 0--2 m. \\
        \addlinespace[3pt]
        Gaussian & $\sigma=0$--$0.15$~m & Add independent Gaussian noise drawn from $\mathcal{N}(0,\sigma^2)$ to each pixel. \\
        \addlinespace[3pt]
        Perlin dropout & 0--20\% of pixels & Drop out pixels using a temporally varying Perlin-noise mask~\cite{perlin2002improving}. \\
        \addlinespace[3pt]
        Vine occlusion & 0--20\% of frames occluded & Place an 8-pixel strip at a random position and orientation; sample its depth uniformly from 0.05--0.20 m. \\
        \midrule
        Pillar hurdle & Hurdle height 0.10--0.40 m & Two pillars beside each hurdle, each three times the hurdle height. \\
        \bottomrule
    \end{tabular}
\end{table}

\begin{table*}[t]

\begin{minipage}[b]{\textwidth}
    \centering
    \caption{Depth-image augmentation and generalization across corruption families.}
    \label{tab:augmentation_transfer}
    \setlength{\tabcolsep}{2pt}
    \renewcommand{\arraystretch}{1.2}
    \footnotesize
    \begin{tabular*}{\textwidth}{@{\extracolsep{\fill}}lcccccccccc@{}}
        \toprule
        \multirow{2}{*}{\textbf{Method}} & \multicolumn{2}{c}{\textbf{Nominal}} & \multicolumn{2}{c}{\textbf{Flying pixels}} & \multicolumn{2}{c}{\textbf{Gaussian}} & \multicolumn{2}{c}{\textbf{Perlin}} & \multicolumn{2}{c}{\textbf{Vine}} \\
        \cmidrule(lr){2-3} \cmidrule(lr){4-5} \cmidrule(lr){6-7} \cmidrule(lr){8-9} \cmidrule(lr){10-11}
        & \textbf{SR} ($\uparrow$) & \textbf{MXD} ($\uparrow$) & \textbf{SR} ($\uparrow$) & \textbf{MXD} ($\uparrow$) & \textbf{SR} ($\uparrow$) & \textbf{MXD} ($\uparrow$) & \textbf{SR} ($\uparrow$) & \textbf{MXD} ($\uparrow$) & \textbf{SR} ($\uparrow$) & \textbf{MXD} ($\uparrow$) \\
        \midrule
ConvNet, no augmentation & $85.9_{\pm 2.3}$ & $0.913$ & $9.3_{\pm 1.9}$ & $0.369$ & $37.3_{\pm 1.6}$ & $0.598$ & $4.4_{\pm 1.8}$ & $0.301$ & $38.1_{\pm 3.6}$ & $0.645$ \\
ConvNet + flying pixels & $91.1_{\pm 3.5}$ & $0.947$ & $87.0_{\pm 3.9}$ & $0.925$ & $84.0_{\pm 2.6}$ & $0.904$ & $13.4_{\pm 2.1}$ & $0.400$ & $46.9_{\pm 5.6}$ & $0.709$ \\
ConvNet + Gaussian & $88.0_{\pm 3.0}$ & $0.933$ & $66.3_{\pm 3.8}$ & $0.814$ & $86.3_{\pm 1.9}$ & $0.923$ & $17.3_{\pm 3.2}$ & $0.431$ & $41.4_{\pm 1.6}$ & $0.674$ \\
ConvNet + Perlin & $77.9_{\pm 4.1}$ & $0.845$ & $35.6_{\pm 4.3}$ & $0.606$ & $72.1_{\pm 4.3}$ & $0.808$ & $69.2_{\pm 3.4}$ & $0.791$ & $41.3_{\pm 3.7}$ & $0.654$ \\
ConvNet + vine occlusion & $88.3_{\pm 2.6}$ & $0.924$ & $1.9_{\pm 0.7}$ & $0.321$ & $25.3_{\pm 2.8}$ & $0.540$ & $3.5_{\pm 0.4}$ & $0.347$ & $83.0_{\pm 2.3}$ & $0.898$ \\
\midrule
\textbf{REDACT, no augmentation} & $96.0_{\pm 1.1}$ & $0.977$ & $91.2_{\pm 2.6}$ & $0.942$ & $93.0_{\pm 1.7}$ & $0.957$ & $47.6_{\pm 4.5}$ & $0.657$ & $65.7_{\pm 2.8}$ & $0.822$ \\
\textbf{REDACT + Perlin} & $95.8_{\pm 1.5}$ & $0.975$ & $75.3_{\pm 2.5}$ & $0.901$ & $\mathbf{94.5}_{\pm 0.7}$ & $\mathbf{0.968}$ & $\mathbf{95.0}_{\pm 1.6}$ & $\mathbf{0.971}$ & $62.4_{\pm 3.2}$ & $0.818$ \\
\textbf{REDACT + vine occlusion} & $\mathbf{96.2}_{\pm 2.2}$ & $\mathbf{0.978}$ & $\mathbf{94.0}_{\pm 1.4}$ & $\mathbf{0.965}$ & $94.3_{\pm 1.6}$ & $0.967$ & $48.4_{\pm 3.6}$ & $0.669$ & $\mathbf{96.2}_{\pm 1.6}$ & $\mathbf{0.978}$ \\
        \bottomrule
    \end{tabular*}
    \par\smallskip
    \begin{minipage}[b]{\textwidth}
    \footnotesize
    \textit{Note:} SR is reported as a percentage and MXD in $[0,1]$. Subscripts give 95\% confidence-interval half-widths; bold values indicate the highest mean in each column. Corruptions are evaluated at the maximum severity in Table~\ref{tab:corruptions}.
    \end{minipage}
    \end{minipage}
\par\bigskip

\begin{minipage}[b]{\textwidth}
    \centering
    \caption{Component ablations under visual corruption and distraction.}
    \label{tab:component_ablation}
    \setlength{\tabcolsep}{2pt}
    \renewcommand{\arraystretch}{1.2}
    \footnotesize
    \begin{tabular*}{\textwidth}{@{\extracolsep{\fill}}lcccccccccccc@{}}
        \toprule
        \multirow{2}{*}{\textbf{Method}} & \multicolumn{2}{c}{\textbf{Nominal}} & \multicolumn{2}{c}{\textbf{Flying pixels}} & \multicolumn{2}{c}{\textbf{Gaussian}} & \multicolumn{2}{c}{\textbf{Perlin}} & \multicolumn{2}{c}{\textbf{Vine}} & \multicolumn{2}{c}{\textbf{Pillar hurdle}} \\
        \cmidrule(lr){2-3} \cmidrule(lr){4-5} \cmidrule(lr){6-7} \cmidrule(lr){8-9} \cmidrule(lr){10-11} \cmidrule(lr){12-13}
        & \textbf{SR} ($\uparrow$) & \textbf{MXD} ($\uparrow$) & \textbf{SR} ($\uparrow$) & \textbf{MXD} ($\uparrow$) & \textbf{SR} ($\uparrow$) & \textbf{MXD} ($\uparrow$) & \textbf{SR} ($\uparrow$) & \textbf{MXD} ($\uparrow$) & \textbf{SR} ($\uparrow$) & \textbf{MXD} ($\uparrow$) & \textbf{SR} ($\uparrow$) & \textbf{MXD} ($\uparrow$) \\
        \midrule
REDACT w/o gate & $\mathbf{96.3}_{\pm 1.4}$ & $\mathbf{0.978}$ & $89.9_{\pm 2.3}$ & $0.936$ & $93.8_{\pm 1.1}$ & $0.961$ & $40.0_{\pm 3.6}$ & $0.599$ & $39.0_{\pm 1.1}$ & $0.663$ & $57.7_{\pm 3.7}$ & $0.746$ \\
REDACT w/o masks & $94.8_{\pm 1.8}$ & $0.970$ & $87.6_{\pm 1.7}$ & $0.923$ & $90.8_{\pm 0.6}$ & $0.943$ & $\mathbf{48.6}_{\pm 2.0}$ & $\mathbf{0.664}$ & $63.9_{\pm 1.7}$ & $0.813$ & $5.1_{\pm 0.6}$ & $0.394$ \\
REDACT w/o masks or gate & $95.4_{\pm 1.3}$ & $0.972$ & $88.3_{\pm 1.4}$ & $0.928$ & $90.6_{\pm 0.9}$ & $0.942$ & $39.0_{\pm 5.1}$ & $0.595$ & $40.5_{\pm 1.4}$ & $0.677$ & $1.9_{\pm 1.0}$ & $0.309$ \\
REDACT w/o GroupNorm & $94.8_{\pm 1.9}$ & $0.968$ & $88.5_{\pm 1.8}$ & $\mathbf{0.950}$ & $\mathbf{94.4}_{\pm 1.3}$ & $\mathbf{0.967}$ & $4.7_{\pm 1.5}$ & $0.313$ & $38.3_{\pm 2.0}$ & $0.663$ & $0.7_{\pm 0.1}$ & $0.335$ \\
REDACT with ConvNet & $84.3_{\pm 2.6}$ & $0.901$ & $32.2_{\pm 1.5}$ & $0.569$ & $61.8_{\pm 1.2}$ & $0.761$ & $5.5_{\pm 1.5}$ & $0.350$ & $55.5_{\pm 2.0}$ & $0.747$ & $2.3_{\pm 0.4}$ & $0.384$ \\
\midrule
\textbf{REDACT (ours)} & $96.0_{\pm 1.1}$ & $0.977$ & $\mathbf{91.2}_{\pm 2.6}$ & $0.942$ & $93.0_{\pm 1.7}$ & $0.957$ & $47.6_{\pm 4.5}$ & $0.657$ & $\mathbf{65.7}_{\pm 2.8}$ & $\mathbf{0.822}$ & $\mathbf{63.8}_{\pm 2.1}$ & $\mathbf{0.792}$ \\
        \bottomrule
    \end{tabular*}
    \par\smallskip
    \begin{minipage}[b]{\textwidth}
    \footnotesize
    \textit{Note:} Metrics, formatting, and corruption severity follow Table~\ref{tab:augmentation_transfer}. Masks denote training-time cell masks. Gated and ungated variants share the same trained encoder and policy; gating is disabled only during evaluation. The ConvNet variant replaces the residual encoder with the ConvNet from~\cite{cheng2023extreme}.
    \end{minipage}
    \end{minipage}
\end{table*}

\subsection{Conformal Calibration and Deployment}
\label{sec:consensus_gate}

We calibrate cell-removal thresholds on clean observations to distinguish nominal prediction disagreement from unusually large deployment deviations. To prevent self-influence, we compare each cell's prediction with the coordinatewise leave-one-out (LOO) median of the other cells:
\begin{equation}
    s_\ell=\left\|\hat u_\ell-\operatorname*{med}_{j\in I\setminus\{\ell\}}\hat u_j\right\|_2.
    \label{eq:consensus_score}
\end{equation}
With at most 11 corrupted cell predictions, each reference has an uncorrupted majority, so its median remains within their coordinatewise range. This reference requires no specified corruption type, but corruption that shifts predictions together may produce little disagreement and remain undetected.

\paragraph{Calibration}
To account for nominal differences between spatial locations, we use approximate conformal calibration~\cite{angelopoulos2023conformal} to establish a separate rejection threshold for each cell using clean rollouts collected by the frozen student with gating disabled. Calibration samples one frame uniformly at random per episode from approximately 3,500 episodes collected offline in Isaac Gym~\cite{makoviychuk2021isaac}, requiring approximately 15 minutes with 192 parallel environments on an NVIDIA L4. Each threshold is the empirical 99th percentile of the cell's disagreement scores, computed with linear interpolation:
\begin{equation}
    \tau_\ell=Q_{0.99}\!\left(\{s_\ell^{(n)}\}_{n=1}^{N_{\mathrm{cal}}}\right),
    \label{eq:calibration_threshold}
\end{equation}
where $n$ indexes the $N_{\mathrm{cal}}$ sampled frames. Exact conformal guarantees require exchangeability between calibration and nominal test scores. Episode-level sampling avoids within-episode temporal dependence, while observation changes under gating motivate treating the empirical thresholds as approximate calibration during deployment.

\paragraph{Deployment}
At each update, cells exceeding their thresholds are excluded from mean--max pooling (Fig.~\ref{fig:redact_overview}(c)). The remaining features are pooled and quantized for the GRU, allowing retained terrain information to support control under partial corruption. If more than 18 cells are flagged, the gate instead removes only the six highest-scoring cells, pooling the remaining 18. This fallback limits feature removal but may retain evidence exceeding its calibrated threshold. Algorithm~\ref{alg:consensus_gate} summarizes the selection and visual-state update.

\raggedbottom
\section{Results}
\label{sec:evaluation}
% We organize the evaluation around four objectives: (1) comparing REDACT's robustness to unseen visual corruption with existing parkour algorithms~\cite{cheng2023extreme,liu2026real}; (2) comparing against augmentation-trained policies and assessing augmentation transfer to unseen noise types; (3) evaluating component contributions through ablation; and (4) assessing zero-shot transfer to indoor, outdoor, and forest terrains with unfamiliar scene content.
We evaluate whether REDACT maintains successful traversal when depth observations differ from those encountered during training. Comparisons with parkour baselines and augmentation-trained policies assess robustness beyond training coverage. We then examine the contributions of the encoder and selective rejection through component ablations and gate analysis. Real-world trials assess zero-shot transfer to structured and forest environments.

\subsection{Experimental Setup}

\textbf{Training and Preprocessing.} REDACT and all baselines use the same terrain generation, curriculum, rewards, and domain randomization following Extreme Parkour~\cite{cheng2023extreme}, controlling for differences in reward shaping and task exposure. We train all policies for a simulated Unitree Go2 in Isaac Gym~\cite{makoviychuk2021isaac} on tilted ramps, gaps, hurdles, and steps (Fig.~\ref{fig:terrains}), using an NVIDIA L4. The shared budget of 15,000 iterations allows all policies to reach convergence in behavioral-cloning loss and the same average terrain-curriculum level. All policies are trained and evaluated with a camera pitched $10^\circ$ downward, rendering depth at $106\times60$ pixels with an $87^\circ$ horizontal field of view. Depth preprocessing consists of cropping dead pixels along the left edge, clipping distances to $[0,2]$ m, and bicubic downsampling to $87\times58$ pixels.

\textbf{Evaluation and Metrics.} We evaluate the four training terrains at uniformly sampled difficulties under Gaussian noise, flying pixels, and Perlin dropout, following prior sensor-error tests and dropout models~\cite{luo2024pie,rudin2025wild}. We additionally test intermittent occlusion from a simulated hanging vine, termed vine occlusion, and visual distraction from pillars placed beside each hurdle without changing the optimal traversal route or required maneuver, termed pillar hurdles (Fig.~\ref{fig:corruptions}). Table~\ref{tab:corruptions} defines each condition and its severity range. Corruption sweeps cover mild degradation through severe stress beyond typical sensor conditions, with image corruptions applied after cropping and before clipping and resizing. Each condition uses 3,200 environments split equally across five seeds. Success rate (SR) measures the proportion of trials reaching all eight randomized checkpoints within 0.2~m of each target. Following~\cite{cheng2023extreme}, mean x-displacement (MXD) records the fraction of checkpoints reached under the same criterion. We report means and 95\% Student-$t$ confidence intervals across five seeds for fixed policies.

\subsection{Comparison to Baselines}
\label{sec:results_baselines}

We compare REDACT with Extreme Parkour~\cite{cheng2023extreme} and REAL~\cite{liu2026real}, retaining the original teacher--student architectures of both baselines and training all three policies without depth-image augmentation. Although all three support locomotion under nominal observations, their success rates diverge under unseen depth corruption (Fig.~\ref{fig:baseline_comparison}). REDACT's higher SR across the tested sweeps suggests that differences in policy design affect robustness beyond the visual conditions encountered during training. This extends to pillar hurdles, where unfamiliar geometry is added beside the route without obscuring the obstacles. Despite retaining the depth information needed for traversal, both baselines exhibited foot-placement and steering errors, suggesting sensitivity to the added visual features rather than missing depth.

\subsection{Generalization Beyond Training Corruptions}

We examine whether depth-image augmentation transfers across corruption families and assess the effects of combining augmentation with REDACT. To evaluate transfer, we train the ConvNet from Extreme Parkour~\cite{cheng2023extreme} separately on each of the four corruption types, then test each policy under all four corruptions and nominal observations. Training and evaluation use the same corruption distributions and application frequencies. We additionally train REDACT with Perlin dropout or vine occlusion, comparing both architectures with their unaugmented counterparts in Table~\ref{tab:augmentation_transfer}. All policies use our regularized multi-query cross-attention teacher to keep teacher supervision consistent. For augmented REDACT policies, GroupNorm scale and shift are learned only from clean depth to limit corruption-specific feature rescaling.

\suppressfloats[t]
\begin{figure}[t]
    \centering
    \includegraphics[width=\linewidth]{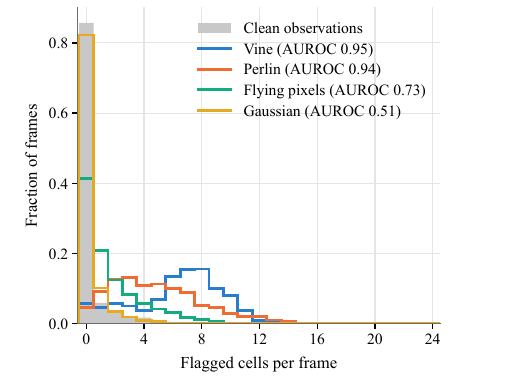}
    \caption{Normalized flagged-cell distributions under clean observations (gray) and four corruptions (step outlines). AUROC values in the legend quantify clean-versus-corrupted frame discrimination using flag count. Corrupted streams use the maximum tested settings.}
    \label{fig:flag_histograms}
    \end{figure}

The ConvNet results show that gains from augmentation extend unevenly to unseen corruption types (Table~\ref{tab:augmentation_transfer}). Flying-pixel and Gaussian augmentation improve performance under both noise types, consistent with their shared effect of perturbing depth values across the image. In contrast, vine-occlusion training provides little benefit under Gaussian noise or Perlin dropout, while flying-pixel SR falls to 1.9\%. Depth-image augmentation therefore does not ensure robustness to corruption types absent from training. Perlin augmentation further reduces nominal ConvNet SR from 85.9\% to 77.9\%, suggesting that extensive depth removal during training can impair nominal performance.

Without depth-image augmentation, REDACT achieves higher SR and MXD than augmentation-trained ConvNet policies under corruptions absent from their training (Table~\ref{tab:augmentation_transfer}). Under flying pixels and Gaussian noise, this advantage persists even against policies trained with the corresponding augmentation, supporting the role of policy design in reducing dependence on training coverage. Targeted augmentation further improves REDACT under Perlin dropout and vine occlusion, although Perlin training also reduces flying-pixel robustness. These findings indicate complementary benefits from policy design for broader robustness and targeted augmentation for specific disturbances.

\subsection{Component Ablations}
\begin{figure}[t]
    \centering
    \includegraphics[width=\linewidth]{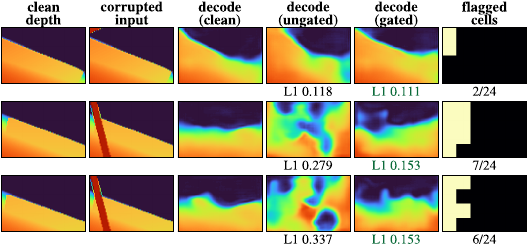}
    \caption{Depth decoding across three frames of simulated vine occlusion. Columns show clean depth, corrupted input, clean-input decoding, ungated and gated corrupted-input decoding, and flagged cells. $L_1$ errors reference the clean-input reconstruction. Flag counts appear below the images.}
    \label{fig:depth_decode_sequence}
    \end{figure}

\label{sec:results_ablation}

Table~\ref{tab:component_ablation} examines encoder architecture, consensus gating, and training-time masking using the same teacher and student objectives. The architectural ablations distinguish contributions from the residual design and feature normalization. Replacing the residual encoder with the ConvNet from~\cite{cheng2023extreme} reduces SR under both nominal observations and all tested disturbances, despite retaining masking and gating. In contrast, removing group normalization from the residual encoder selectively reduces SR under Perlin dropout, vine occlusion, and pillar hurdles, while nominal traversal and performance under pixelwise noise remain high. This pattern is consistent with normalization limiting sensitivity to altered feature scales under missing depth and unfamiliar geometry.

To isolate the effect of feature rejection, we evaluate the same trained checkpoint with gating enabled and disabled. Enabling gating leaves nominal SR largely unchanged while raising vine-occlusion SR from 39.0\% to 65.7\%, with further gains under Perlin dropout and pillar hurdles. These gains support the role of gating in preserving useful evidence when disturbances affect spatially concentrated features. Training-time masking further supports traversal under partial evidence, with its removal sharply reducing pillar-hurdle SR regardless of gating. Its benefit therefore extends beyond preparing the controller for information removed by the gate.

\subsection{Gate Analysis}
\label{sec:results_detection}

We examine how consensus flag counts respond to different visual corruptions. Fig.~\ref{fig:flag_histograms} shows distributions collected across 50,000 frames in total, with corrupted distributions restricted to frames in which corruption is present. Perlin dropout and vine occlusion produce higher flagged-cell counts than clean observations, consistent with spatially concentrated corruption exceeding the clean-calibrated disagreement thresholds. This separation yields areas under the receiver operating characteristic curve (AUROC) of 0.94 for Perlin dropout and 0.95 for vine occlusion, using flag count to distinguish corrupted from clean frames. In contrast, flying pixels show weaker discrimination (AUROC 0.73), while Gaussian noise produces near-chance discrimination (AUROC 0.51). Spatially distributed perturbations may affect both individual predictions and their LOO median reference, limiting detection. Nevertheless, Table~\ref{tab:component_ablation} shows that REDACT maintains high traversal success under Gaussian noise with gating disabled, indicating that robustness in this condition does not depend on selective rejection.

Beyond frame-level discrimination, we examine whether gating preserves the pooled representation using a diagnostic decoder trained on clean depth frames. The decoder reverses the residual encoder architecture in Fig.~\ref{fig:residual_encoder}, is trained separately from the policy, and is applied to both gated and ungated latents. In the vine-occlusion sequence in Fig.~\ref{fig:depth_decode_sequence}, gated reconstructions more closely resemble the corresponding clean-input reconstructions and have lower $L_1$ error against that reference. The flagged cells align spatially with the approximate location of the occluder.

\subsection{Real-World Traversal}
\begin{figure}[t]
    \centering
    \includegraphics[width=\columnwidth]{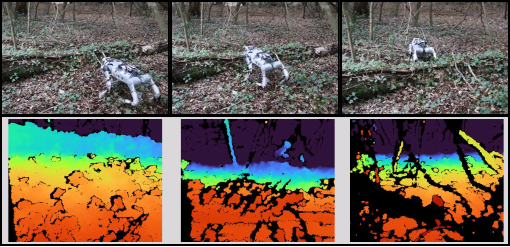}
    \caption{Real-world log traversal amid vegetation and leaf litter. External views (top) show successive traversal stages; cropped onboard depth (bottom) shows dropouts from vegetation and additional tree and branch geometry. Frames are drawn from the same trial in traversal order.}
    \label{fig:forest_depth_sequence}
\end{figure}

We evaluate real-world deployment on a Unitree Go2 equipped with an NVIDIA Jetson Orin Nano and an Intel RealSense D435i depth camera. Repeated trials in the settings in Fig.~\ref{fig:real_world_teaser} assess visual distraction from pillars beside hurdles (pillar hurdles) and fallen timber perpendicular to a traversal log (log + distractors). Hurdles without pillars provide a nominal reference, while a second log setting combines vegetation-induced depth dropouts with partial occlusion by a leaning tree (log + vegetation). A representative traversal with cropped onboard depth images is shown in Fig.~\ref{fig:forest_depth_sequence}. The depth backbone operates at 10 Hz and the MLP actor at 50 Hz. Median visual-update latency is 8.8~ms with gating, 6.2~ms without gating, and 1.9~ms with the ConvNet~\cite{cheng2023extreme}.

Table~\ref{tab:real_world} reports real-world results for the policies from Section~\ref{sec:results_baselines}, all of which successfully traverse nominal hurdles. With adjacent pillars, we observed the baselines steering away from the intended route, often colliding with the cardboard pillars before the trials were stopped. In comparison, REDACT showed little change in gait quality with the added distractors. Given the observed collisions, baseline evaluations were not extended to the logs to protect the hardware. In the forest settings, REDACT completed seven of ten trials at each site. Although the logs resemble the hurdle obstacles used during training, their smooth, curved surfaces introduced slipping. Roots in the vegetated setting introduced additional tripping hazards, with some trips followed by recovery. Additional outdoor demonstrations show successful traversal of hurdles, steps, and gravel under sunny and low ambient light, with representative examples in Fig.~\ref{fig:real_world_teaser}.

% \begin{figure}[H]
%     \centering
%     \includegraphics[width=\columnwidth]{Figures/real_world_dark.pdf}
%     \caption{Real-world demonstration of successful traversal over hurdles, steps, and gravel terrain under low ambient lighting.}
%     \label{fig:real_world_dark}
% \end{figure}

\begin{table}[H]
    \centering
    \caption{Real-world traversal: successes/trials (settings in Fig.~\ref{fig:real_world_teaser}). Dashes indicate settings not evaluated for hardware safety.}
    \label{tab:real_world}
    \renewcommand{\arraystretch}{1.15}
    \setlength{\tabcolsep}{2pt}
    \footnotesize
    \begin{tabular*}{\columnwidth}{@{\extracolsep{\fill}}lcccc@{}}
        \toprule
        \textbf{Method} & \textbf{Hurdles} & \shortstack{\textbf{Pillar}\\\textbf{hurdles}} & \shortstack{\textbf{Log +}\\\textbf{distractors}} & \shortstack{\textbf{Log +}\\\textbf{vegetation}} \\
        \midrule
        REDACT (ours) & 5 / 5 & 5 / 5 & 7 / 10 & 7 / 10 \\
        REAL~\cite{liu2026real} & 5 / 5 & 0 / 5 & -- & -- \\
        Extreme Parkour~\cite{cheng2023extreme} & 4 / 5 & 0 / 5 & -- & -- \\
        \bottomrule
    \end{tabular*}
\end{table}

\section{Conclusion}

In this work, we propose REDACT, a framework for retaining terrain information under visual degradation absent from training. REDACT combines an improved visual encoder architecture and persistent feature masking with a novel consensus-gating mechanism that excludes inconsistent spatial features before aggregation. Our evaluations demonstrate improved robustness to unseen disturbances, with ablations showing contributions from encoder design, masking, and selective rejection. Comparisons with augmentation-trained policies show that REDACT supports robustness where training coverage is missing, while benefiting from targeted depth augmentation. Real-world trials demonstrate zero-shot traversal in structured and forest environments with unfamiliar scene content.

\section*{Acknowledgment}
ChatGPT was used for LaTeX formatting of the equations in Section III and linguistic refinement in Sections II and V. The authors reviewed and revised all technical content, results, and text to ensure consistency with the research findings and academic standards. 

%The authors also acknowledge the use of the IRIDIS X High Performance Computing Facility and the Southampton-Wolfson AI Research Machine (SWARM) GPU cluster, funded by the Wolfson Foundation.

\FloatBarrier
\bibliographystyle{IEEEtran}
\bibliography{references}

\end{document}